\documentclass[letterpaper]{article} 
\usepackage[preprint]{aaai2027}  
\usepackage[hyphens]{url}  
\usepackage{graphicx} 
\usepackage{natbib}  
\usepackage{caption} 
\usepackage{algorithm}
\usepackage{algorithmic}
\usepackage{amsmath}
\usepackage{booktabs}
\usepackage{multirow}
\usepackage{newfloat}
\usepackage{listings}
\DeclareCaptionStyle{ruled}{labelfont=normalfont,labelsep=colon,strut=off} 
\floatstyle{ruled}
\newfloat{listing}{tb}{lst}{}
\floatname{listing}{Listing}

\usepackage{booktabs}

\title{SkillIR: Evolving Scene-Aware Skills for Agentic Image Restoration}
\title{SkillIR: Evolving Scene-Aware Skills for Agentic Image Restoration}

\author{
    Jie Shao\textsuperscript{\rm 1},
    Shengkai Hu\textsuperscript{\rm 1},
    Xu Zhang\textsuperscript{\rm 2},
    Beihang Song\textsuperscript{\rm 4},
    Yongcheng Jing\textsuperscript{\rm 2},
    Xu Wu\textsuperscript{\rm 3},
    Jun Wan\textsuperscript{\rm 1}\corresponding
}

\affiliations{
    \textsuperscript{\rm 1}Zhongnan University of Economics and Law, Wuhan, China\\
    \textsuperscript{\rm 2}Wuhan University, Wuhan, China\\
    \textsuperscript{\rm 3}Dongguan University of Technology, Dongguan, China\\
    \textsuperscript{\rm 4}National Institute of Natural Hazards, Ministry of Emergency Management of China, Beijing, China
}

\begin{document}

\maketitle


\begin{abstract}
This paper studies agentic image restoration, in which multimodal agents coordinate specialized restoration tools to recover images affected by complex degradations. Existing restoration agents often derive complete tool-use plans from the original degraded image or retrieve previously successful trajectories, providing limited support for adapting individual actions to evolving intermediate restoration states. We find that accepted tool executions can change the residual degradation state and, consequently, the applicability of subsequent tools.
To address this issue, we propose \textbf{SkillIR}, a skill-guided framework that represents restoration experience as degradation-centered action evidence rather than complete tool-use trajectories. SkillIR consolidates context-dependent action outcomes into scene-aware restoration skills that characterize applicable conditions, expected effects, and attributable failure cases. Instead of prescribing a complete restoration plan, the retrieved skills guide one bounded action at a time within a verified residual-state loop: each tool output is treated as a candidate, committed only after transition verification, and followed by reassessment of the active residual degradations. After each rollout, the resulting evidence is used to create, refine, or patch dynamic skills, enabling accumulated restoration experience to improve decision-making for subsequent inputs.
Experiments on synthetic and real-world multi-degradation datasets demonstrate that SkillIR improves restoration quality and enables more reliable and effective tool use.
\end{abstract}

\section{Introduction}

Image restoration is a fundamental task in computer vision that aims to recover high-quality images from degraded observations~\cite{sun2022rethinking,yang2023visual}. Existing task-specific methods have achieved remarkable success in individual restoration tasks, including denoising~\cite{chen2022simple}, deraining~\cite{dong2025channel}, dehazing~\cite{song2023vision}, deblurring~\cite{ruan2022learning}, and low-light enhancement~\cite{lv2024fourier}. However, real-world images frequently contain multiple coupled degradations whose interactions cannot be fully captured by isolated task formulations. Although All-in-One image restoration (AiOIR) methods~\cite{PromptIR,conde2024instructir} unify multiple degradation-specific mappings within a single network, they remain limited in handling out-of-distribution degradation compositions and dynamically changing restoration states.

\begin{figure*}[t]
\centering
\includegraphics[width=0.96\linewidth]{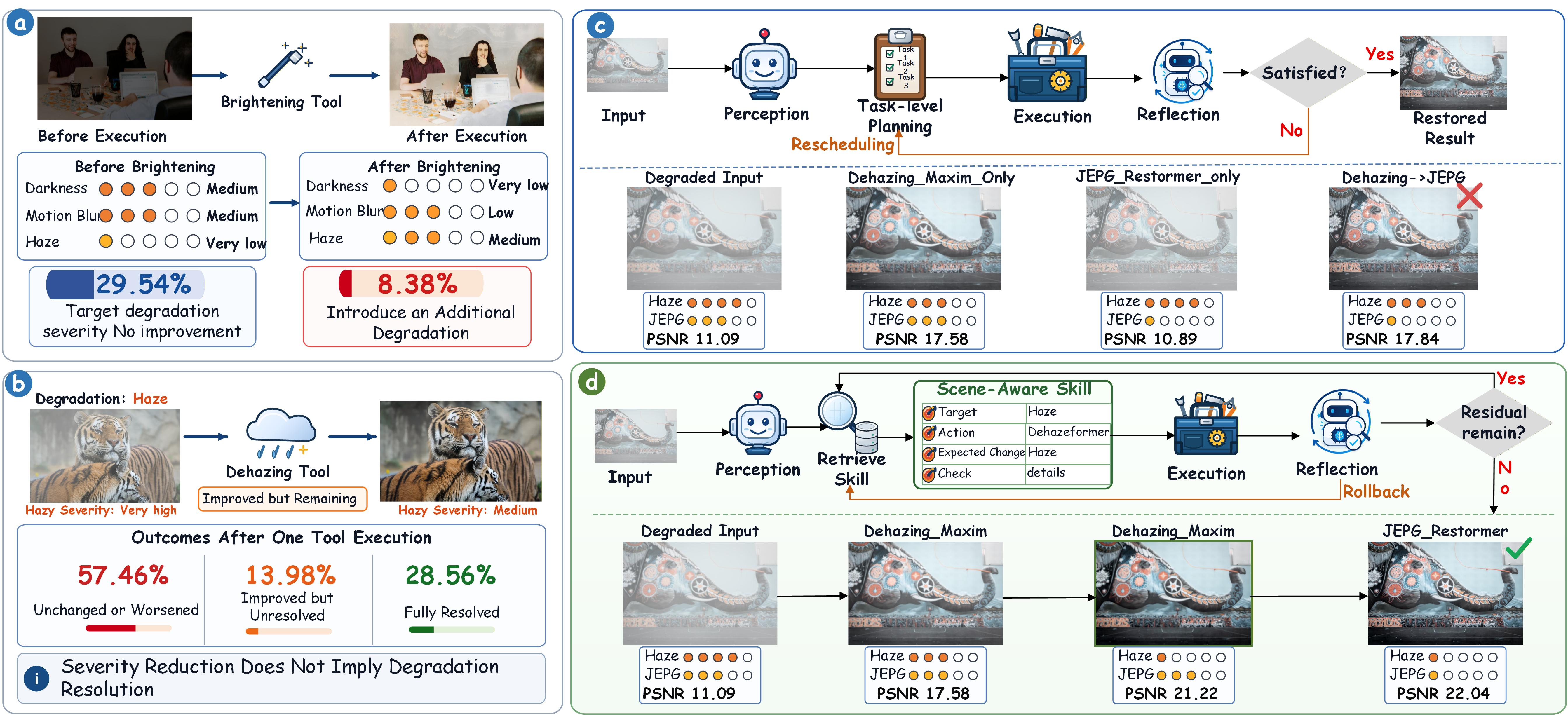}
\vspace{-0.5em}
\caption{
Motivation and conceptual comparison of SkillIR. (a) Cross-degradation side effects: executing a restoration action (e.g., brightening) may inadvertently introduce or exacerbate other degradations. (b) Action-level progress vs. degradation-level resolution: an accepted step often only mitigates degradation severity rather than eliminating it, leaving residual artifacts. (c) Conventional agent architectures perform task-level tool orchestration without fine-grained state verification, frequently leading to premature termination or sub-optimal plans. (d) SkillIR addresses these challenges by incorporating scene-aware skill retrieval and verifying residual states after each action, enabling flexible rollback and multi-step iterative restoration until full resolution.}
\label{fig:intro_model}
\vspace{-1.5em}
\end{figure*}


Recent advances in Multimodal Large Language Models (MLLMs) have enabled agentic image restoration, in which MLLM-based agents determine the order in which degradations are addressed and sequentially invoke the corresponding specialist restoration tools~\cite{chen2024restoreagent,zhu2025agenticir}. Existing approaches can be broadly grouped according to how tool-use experience is encoded. Policy-learning methods internalize tool selection and ordering knowledge through reinforcement learning or agentic post-training~\cite{lu2026restorer1,zhang2026tiragent}, enabling efficient inference but typically requiring additional optimization to accommodate new tools or degradation types. In contrast, training-free methods keep the controller fixed and reuse explicit experience through execution records, reflection, or trajectory memory~\cite{zhu2025agenticir,jiang2025mair,zuo20254kagent,cui2026sear}, offering greater flexibility but commonly retaining experience at the level of complete plans or trajectories.


Although training-free agents can revise their decisions through re-perception, reflection, or quality feedback, every accepted action changes the current restoration state and may alter both the active residual degradations and the actions that are suitable next. As illustrated in Fig.~\ref{fig:intro_model}(a)--(b), a successful action may reduce the degradation being processed without fully resolving it, and may also affect co-existing degradations or previously restored content. Therefore, an improved result does not necessarily mean that the current degradation has been resolved or that the previous restoration decision remains appropriate. Switching to another degradation too early may leave the current one unresolved, whereas repeating an action that is no longer suitable may introduce artifacts or damage restored details.

In this work, we propose \textbf{SkillIR}, a framework that equips image restoration agents with reusable scene-aware skills. SkillIR learns these skills by decomposing tool-use trajectories into degradation-centered episodes and aggregating compatible verified transitions, while retaining rejected attempts as failure lessons that characterize action risks and applicability boundaries. Critically, each skill provides state-conditioned guidance for only the next bounded action rather than prescribing a complete restoration trajectory. Transition verification and residual re-perception then keep subsequent decisions aligned with the updated intermediate state, while newly verified outcomes from each completed rollout further refine the skills available to later inputs. As summarized in Fig.~\ref{fig:intro_model}(c)--(d), this design replaces task-level trajectory guidance with state-adaptive, verified action selection.

The main contributions are summarized as follows:
\begin{itemize}

\item We propose \textbf{SkillIR}, an agentic image restoration framework that decomposes completed tool-use trajectories into degradation-centered episodes and distills verified local transitions into scene-aware skills and failure lessons. This representation enables state-conditioned reuse of action knowledge as intermediate restoration states evolve.

\item We develop a skill-guided verified restoration loop that combines bounded action selection, transition verification, and residual re-perception. By keeping each decision aligned with the updated residual state, SkillIR supports adaptive continuation, task switching, and termination while reducing premature decisions and state-mismatched tool use.

\item Experiments on synthetic and real-world multi-degradation datasets demonstrate that SkillIR delivers superior restoration performance while optimizing tool execution.
\end{itemize}

\begin{figure*}[t]
\centering
\includegraphics[width=0.96\linewidth]{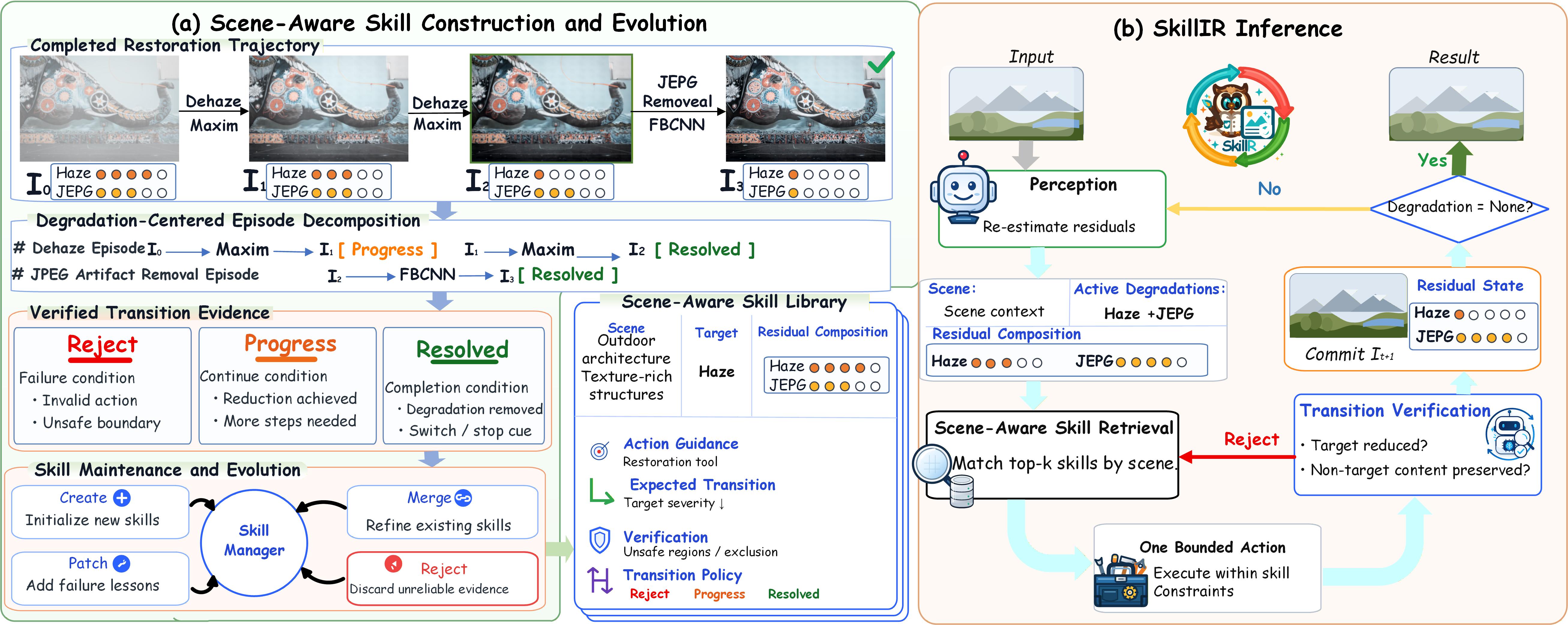}
\vspace{-0.5em}
\caption{Architectural overview of SkillIR.
(a) Scene-Aware Skill Construction and Evolution: SkillIR decomposes verified restoration trajectories into degradation-centered episodes and consolidates them into scene-aware skills, while retaining rejected transitions as failure lessons. Dynamic skills are updated using verified evidence from completed rollouts.
(b) Skill-Guided Verified Restoration: SkillIR retrieves skills matched to the current scene and residual state to guide one bounded action at a time, verifies each candidate before commitment, and re-perceives residual degradations for subsequent decisions.}
\label{fig:framework}
\vspace{-1.5em}
\end{figure*}

\section{Related work}
\subsection{All-in-One Image Restoration}

All-in-One Image Restoration (AiOIR) aims to handle multiple degradation types within a unified model. Existing methods primarily improve unified restoration through degradation-aware representations, adaptive capacity allocation, and broader generalization. Prompt-based approaches encode degradation characteristics to guide restoration across different types and severities~\cite{PromptIR,ProRes,perceiveir}, while expert-routing and feature-adaptation methods alleviate task interference by allocating model capacity according to the input degradation~\cite{moceir2025,dfpir2025}. Other studies extend AiOIR to multi-domain or generative settings through domain-specific prompts, autoregressive modeling, or semantic degradation assessment~\cite{datprlir,restorevar,clearair}. However, their generalization to out-of-distribution degradation combinations remains limited, especially in intricate real-world scenarios that require adaptive and sequential restoration decisions.

\subsection{Agentic Image Restoration}

Agentic image restoration extends conventional restoration pipelines by using multimodal agents to identify degradations, select specialized tools, and assess intermediate results. Existing approaches can be broadly divided into training-free and training-based methods according to how tool-use knowledge is acquired and reused.
\emph{Training-free methods} keep the controller parameters fixed and guide restoration through prompting, multi-agent coordination, iterative reflection, search, or external memory~\cite{chen2024restoreagent,zhu2025agenticir,jiang2025mair,li2026hybridagent,zhou2025qagent,zuo20254kagent,cui2026sear}. These methods can flexibly incorporate new restoration tools without additional model training. However, their experience is often used only within the current rollout or retained as complete plans and trajectories, which may lead to repeated trial-and-error when the restoration state changes.
\emph{Training-based methods} optimize restoration decisions through reinforcement learning or agentic post-training~\cite{lu2026restorer1,zhang2026tiragent,zhu2026opera}. By internalizing tool selection and planning strategies in model parameters, they can produce more consistent and efficient decisions, but typically require additional optimization when new tools or degradation types are introduced.

Despite their different forms, existing methods generally represent restoration outcomes at a relatively coarse level. In particular, they rarely distinguish partial degradation reduction from complete resolution as explicit verified transitions that can guide both subsequent actions and experience reuse. 



\section{Method}

As shown in Fig.~\ref{fig:framework}, we propose SkillIR, a skill-guided framework for agentic image restoration. SkillIR comprises two complementary components: scene-aware skill construction and evolution and skill-guided verified restoration. The former abstracts verified tool-use trajectories into degradation-centered restoration skills and failure lessons, while the latter uses the acquired skills to guide bounded restoration actions through transition verification and residual-state updates.

\subsection{Scene-Aware Skill Construction and Evolution}

SkillIR converts verified tool-use trajectories into reusable, state-dependent restoration evidence. Since the effect of a restoration action depends not only on the degradation being processed but also on the scene content and co-existing residual degradations, SkillIR first organizes trajectory evidence at the degradation level and then consolidates it into structured skills and failure lessons. The resulting dynamic skills are progressively refined using verified evidence from completed rollouts.

\noindent\textbf{Degradation-Centered Trajectory Decomposition.}
A complete restoration trajectory often interleaves decisions made for multiple degradations, making it difficult to attribute an action outcome to a specific restoration objective. SkillIR therefore decomposes each trajectory into degradation-centered episodes. Each episode is defined as a maximal contiguous subsequence of decisions and verified transitions associated with one degradation being processed. It begins when that degradation is selected and ends when it is resolved, deferred, or the agent switches to another active degradation; revisiting the same degradation later initiates a new episode. By preserving the corresponding Progress, Resolved, and Rejected transitions, each episode provides localized and attributable evidence for subsequent skill construction and revision.

\noindent\textbf{Restoration Skill Construction.}
Given a degradation-centered episode, SkillIR abstracts its verified execution evidence into a state-dependent restoration skill rather than preserving it as a rigid tool sequence. Each skill specifies its applicability conditions, including the scene context, the degradation being processed, and the co-existing residual degradations, together with admissible actions and parameter bounds, expected restoration effects, and verification criteria. A standardized MLLM prompt converts the episode evidence into this structured representation by jointly considering the restoration context and verified transition outcomes. A fixed model and decoding configuration are used throughout skill construction, with full prompt details provided in the Technical Supplement.

\noindent\textbf{Failures as Lessons.}
Rejected transitions are retained as failure lessons when their outcomes can be reliably attributed to the attempted actions. Each lesson records the restoration context, attempted action, and observed failure mode, such as insufficient degradation removal, structural damage, artifact introduction, or tool--context mismatch. It further specifies applicable exclusion conditions or fallback guidance, helping the agent avoid repeating harmful or sub-optimal actions in analogous restoration states.

\noindent\textbf{Skill Maintenance and Evolution.}
The Skill Manager incorporates verified episode evidence through four maintenance operations. \emph{Create} initializes a dynamic skill when valid positive evidence is not covered by existing skills; \emph{Merge} integrates compatible \emph{Progress} or \emph{Resolved} evidence to refine applicability conditions, action constraints, and expected outcomes; \emph{Patch} attaches attributable \emph{Rejected} evidence as failure lessons and updates the exclusion or fallback rules; and \emph{Reject} discards incomplete, inconsistent, or insufficiently verified evidence. The maintenance operation for each episode is determined by the Skill Manager using a fixed MLLM prompt and decoding configuration. The structured skill schema, decision criteria, and prompt template are provided in the supplementary material.
Static skills remain unchanged, whereas dynamic skills evolve only through validated updates. The initial Skill Library is constructed from verified exploration trajectories. During deployment, the library remains fixed throughout the restoration of each individual image and is updated only after the rollout is completed. Consequently, the resulting dynamic skills are used exclusively for subsequent inputs. These updates rely only on tool-use observations and transition-verification outcomes, without access to reference images or ground-truth degradation labels.

\begin{table*}[t]
\centering
\small
\setlength{\tabcolsep}{2.8pt}
\renewcommand{\arraystretch}{1.12}
\resizebox{0.92\textwidth}{!}{%
\begin{tabular}{@{}lclccccccc@{}}
\toprule
Dataset & Type & Method & Venue
& PSNR$\uparrow$
& SSIM$\uparrow$
& LPIPS$\downarrow$
& MANIQA$\uparrow$
& CLIP-IQA$\uparrow$
& MUSIQ$\uparrow$ \\
\midrule

\multirow{11}{*}{Group A}
& \multirow{6}{*}{All-in-One}
& AirNet~\cite{li2022airnet}
& CVPR'22
& 19.13 & 0.6019 & 0.4283 & 0.2581 & 0.39300 & 42.46 \\

& & PromptIR~\cite{PromptIR}
& NeurIPS'23
& 20.06 & 0.6088 & 0.4127 & 0.2633 & 0.40130 & 42.62 \\

& & MiOIR~\cite{kong2024mioir}
& arXiv'24
& 20.84 & 0.6558 & 0.3715 & 0.2451 & 0.39330 & 47.82 \\

& & DA-CLIP~\cite{luo2024daclip}
& ICLR'24
& 19.58 & 0.6032 & 0.4266 & 0.2418 & 0.41390 & 42.51 \\

& & InstructIR~\cite{conde2024instructir}
& ECCV'24
& 18.03 & 0.5751 & 0.4429 & 0.2660 & 0.35280 & 45.77 \\

& & AutoDIR~\cite{jiang2024automatic}
& ECCV'24
& 19.64 & 0.6286 & 0.3967 & 0.2500 & 0.37670 & 47.01 \\

\cmidrule(lr){2-10}

& \multirow{5}{*}{Agent}
& AgenticIR~\cite{zhu2025agenticir}
& ICLR'25
& 21.04 & 0.6818 & 0.3148 & 0.3071 & 0.44740 & 56.88 \\

& & MAIR~\cite{jiang2025mair}
& IJCV'26
& 21.02 & 0.6715 & \underline{0.2963}
& 0.3330 & 0.47510 & 59.19 \\

& & 4KAgent~\cite{zuo20254kagent}
& NeurIPS'25
& 21.48 & 0.6720 & 0.3019
& \textbf{0.3748} & \textbf{0.55440} & \textbf{63.19} \\

& & SEAR~\cite{cui2026sear}
& ECCV'26
& \underline{21.80} & \underline{0.6961} & 0.3019
& 0.3411 & 0.50390 & 61.29 \\

& & \textbf{Ours}
& --
& \textbf{22.39} & \textbf{0.7145} & \textbf{0.2744}
& \underline{0.3620} & \underline{0.52881} & \underline{62.31} \\

\midrule

\multirow{11}{*}{Group B}
& \multirow{6}{*}{All-in-One}
& AirNet~\cite{li2022airnet}
& CVPR'22
& 19.31 & 0.6567 & 0.3670 & 0.2882 & 0.42740 & 47.88 \\

& & PromptIR~\cite{PromptIR}
& NeurIPS'23
& 20.47 & 0.6704 & 0.3370 & 0.2893 & 0.42890 & 48.10 \\

& & MiOIR~\cite{kong2024mioir}
& arXiv'24
& 20.56 & 0.6905 & 0.3243 & 0.2638 & 0.43300 & 51.87 \\

& & DA-CLIP~\cite{luo2024daclip}
& ICLR'24
& 18.56 & 0.5946 & 0.4405 & 0.2435 & 0.41540 & 43.70 \\

& & InstructIR~\cite{conde2024instructir}
& ECCV'24
& 18.34 & 0.6235 & 0.4072 & 0.3022 & 0.37900 & 50.94 \\

& & AutoDIR~\cite{jiang2024automatic}
& ECCV'24
& 19.90 & 0.6643 & 0.3542 & 0.2534 & 0.39860 & 49.64 \\

\cmidrule(lr){2-10}

& \multirow{5}{*}{Agent}
& AgenticIR~\cite{zhu2025agenticir}
& ICLR'25
& 20.55 & 0.7009 & 0.3072 & 0.3204 & 0.46480 & 57.57 \\

& & MAIR~\cite{jiang2025mair}
& IJCV'26
& 20.92 & 0.7004 & \underline{0.2788}
& 0.3544 & 0.50840 & 60.98 \\

& & 4KAgent~\cite{zuo20254kagent}
& NeurIPS'25
& 20.95 & 0.6727 & 0.3017
& \textbf{0.3734} & \textbf{0.55050} & \textbf{62.69} \\

& & SEAR~\cite{cui2026sear}
& ECCV'26
& \textbf{22.13} & \textbf{0.7251} & 0.2890
& 0.3447 & 0.51340 & 60.65 \\

& & \textbf{Ours}
& --
& \underline{21.73} & \underline{0.7228} & \textbf{0.2655}
& \underline{0.3675} & \underline{0.55046} & \underline{62.60} \\

\midrule

\multirow{11}{*}{Group C}
& \multirow{6}{*}{All-in-One}
& AirNet~\cite{li2022airnet}
& CVPR'22
& 17.95 & 0.5145 & 0.5782 & 0.1854 & 0.31130 & 30.12 \\

& & PromptIR~\cite{PromptIR}
& NeurIPS'23
& 18.51 & 0.5166 & 0.5756 & 0.1906 & 0.31040 & 29.71 \\

& & MiOIR~\cite{kong2024mioir}
& arXiv'24
& 15.63 & 0.4896 & 0.5376 & 0.1717 & 0.28910 & 37.95 \\

& & DA-CLIP~\cite{luo2024daclip}
& ICLR'24
& 18.53 & 0.5320 & 0.5335 & 0.1916 & 0.34760 & 33.87 \\

& & InstructIR~\cite{conde2024instructir}
& ECCV'24
& 17.09 & 0.5135 & 0.5582 & 0.1732 & 0.25370 & 33.69 \\

& & AutoDIR~\cite{jiang2024automatic}
& ECCV'24
& 18.61 & 0.5443 & 0.5019 & 0.2045 & 0.29390 & 37.86 \\

\cmidrule(lr){2-10}

& \multirow{5}{*}{Agent}
& AgenticIR~\cite{zhu2025agenticir}
& ICLR'25
& 18.82 & 0.5474 & 0.4493 & 0.2698 & 0.39480 & 48.68 \\

& & MAIR~\cite{jiang2025mair}
& IJCV'26
& 19.42 & 0.5544 & \underline{0.4142}
& 0.2798 & 0.42390 & 51.36 \\

& & 4KAgent~\cite{zuo20254kagent}
& NeurIPS'25
& 19.77 & 0.5629 & 0.4271
& \textbf{0.3545} & \textbf{0.52330} & \textbf{55.56} \\

& & SEAR~\cite{cui2026sear}
& ECCV'26
& \textbf{20.61} & \underline{0.5927} & 0.4173
& 0.2742 & 0.44200 & 52.73 \\

& & \textbf{Ours}
& --
& \underline{20.59} & \textbf{0.6039} & \textbf{0.4031}
& \underline{0.3047} & \underline{0.45033} & \underline{54.57} \\

\bottomrule
\end{tabular}%
}
\vspace{-0.9em}
\caption{Quantitative comparison with previous state-of-the-art
methods on the three multiple-degradation subsets of the MiO100
dataset. The best and second-best results for each metric are
marked in bold and underline, respectively.}
\label{tab:mio100_all_merged_new}
\vspace{-1.5em}
\end{table*}

\subsection{Skill-Guided Verified Restoration}

Given the available Skill Library, SkillIR restores each input through a verified residual-state loop. Rather than committing to a complete tool sequence, it retrieves state-matched skills to guide one specialist-tool execution at a time. Since each execution may affect both the degradation being processed and other image content, its output is treated as a candidate, verified before commitment, and followed by residual re-perception before the next decision.

For the $n$-th input image, let $I_t$ denote the latest committed image before the $t$-th action attempt. SkillIR represents the current restoration context as
\begin{equation}
\mathbf c_t=(\mathbf s,d_t,\mathbf z_t,\mathcal H_t),
\end{equation}
where $\mathbf s$ describes the scene context, $d_t$ denotes the active degradation selected for processing, $\mathbf z_t$ records the active residual degradations together with their estimated severity and confidence, and $\mathcal H_t$ contains previous action attempts and verified transition outcomes within the current rollout.

\noindent\textbf{Scene-Aware Skill Retrieval.}
The same degradation may require different actions across scenes or in the presence of different residual degradations. SkillIR therefore retrieves guidance according to the full restoration context rather than matching only the degradation label. For the $n$-th image, the available library is
$\mathcal{L}^{(n)}=\mathcal{L}^{\mathrm{sta}}\cup\mathcal{L}^{\mathrm{dyn},(n)}$,
where $\mathcal{L}^{\mathrm{sta}}$ contains fixed skills initialized from verified exploration trajectories, and
$\mathcal{L}^{\mathrm{dyn},(n)}$ contains dynamic skills created or updated from completed rollouts preceding the $n$-th image. The library remains fixed throughout the current rollout and is updated only after the rollout is completed.
Given $\mathbf c_t$, the MLLM ranks the available skills according to their compatibility with the scene, the degradation being processed, the active residual degradations, and previous restoration outcomes. The top-$k$ retrieved skills, denoted by $\mathcal K_t$, provide action guidance and applicable failure lessons for the current state. If no retrieved skill is sufficiently compatible with $\mathbf c_t$, the controller falls back to constrained direct planning over the restoration toolbox, subject to applicable failure lessons, actions rejected earlier in the current rollout, and execution constraints. The resulting transition is still verified and may provide evidence for skill creation after the rollout is completed.

\noindent\textbf{Skill-Guided Bounded Action Selection.}
Because the restoration state may change after every tool execution, committing to a complete tool sequence in advance can make later actions inappropriate. SkillIR instead selects one bounded restoration action at each step, where a bounded action corresponds to a single specialist-tool invocation. Conditioned on $\mathbf c_t$ and $\mathcal K_t$, the controller resolves overlapping or conflicting guidance and selects a specialist restoration tool $f_t$. The selected action is admissible only if it is supported by the retrieved skills or generated through the constrained fallback, and is not excluded by applicable failure lessons, actions rejected earlier in the current rollout, or execution constraints. The next action is selected only after the effect of the current action has been observed.

\noindent\textbf{Verified Transition Assessment.}
A specialist tool may fail to sufficiently reduce the intended degradation or may damage other image content. Its output is therefore treated as a candidate rather than being committed directly. Executing $a_t$ produces
$\widetilde{I}_{t+1}=f_t(I_t)$. The candidate is accepted only when the transition verifier determines that it reliably reduces $d_t$ without causing unacceptable damage elsewhere. The committed image is updated as
\begin{equation}
I_{t+1}=
\begin{cases}
\widetilde{I}_{t+1}, & \text{if the candidate is accepted},\\
I_t, & \text{otherwise}.
\end{cases}
\end{equation}
If the candidate is rejected, it is discarded and the transition is recorded as \emph{Rejected}. Since an accepted action does not necessarily eliminate the degradation completely, SkillIR re-perceives $I_{t+1}$ to update the residual degradation state. The transition is recorded as \emph{Progress} if $d_t$ remains active, or as \emph{Resolved} if $d_t$ is no longer active. The attempted action and transition outcome are then appended to $\mathcal H_{t+1}$.

\noindent\textbf{Residual-State Update and Scheduling.}
The verified transition and updated residual state determine the next restoration decision. After a \emph{Rejected} transition, SkillIR retrieves skills again under the updated history while excluding the rejected action from the current rollout context. After a \emph{Progress} transition, $d_t$ remains the current target, and SkillIR performs a new retrieval and action-selection step if another admissible action is available. After a \emph{Resolved} transition, the scheduler selects another active degradation.
If $d_t$ remains active but no admissible action is available under the current context, it is temporarily deferred so that another active degradation can be processed. The deferred degradation may be revisited after the residual state changes. The process terminates when no processable active degradation remains or when the execution budget is exhausted, and the latest committed image is returned as the final result.

\section{Experiments}

\subsection{Experimental Setup}

\noindent\textbf{Implementation Details.} We follow the experimental setting of SEAR~\cite{cui2026sear}. The restoration toolbox consists of pretrained X-Restormer~\cite{chen2024comparative}, HAT~\cite{chen2023activating}, FBCNN~\cite{jiang2021towards}, SwinIR~\cite{liang2021swinir}, DiffBIR~\cite{lin2024diffbir}, MPRNet~\cite{zamir2021multi}, DRBNet~\cite{ruan2022learning}, DehazeFormer~\cite{song2023vision}, MAXIM~\cite{tu2022maxim}, RIDCP~\cite{wu2023ridcp}, Restormer~\cite{zamir2022restormer}, and several traditional image-processing operators. Following AgenticIR~\cite{zhu2025agenticir}, we fine-tune two DepictQA checkpoints~\cite{you2024depicting} for residual-degradation perception and before--after transition verification, respectively. 
GPT-4o~\cite{openai2024gpt4o} serves as the high-level controller for degradation selection, skill retrieval, and bounded action planning, with the top-$3$ retrieved skills provided at each decision step. The initial Skill Library contains static skills distilled from verified exploration episodes collected on the MiO100 training split. 
During deployment, these static skills remain fixed, while dynamic skills are progressively constructed and updated from verified episodes obtained from completed image rollouts. The library remains unchanged within each individual rollout, so evidence from the current image can influence only subsequent inputs. All experiments are conducted on two NVIDIA RTX 4090 GPUs. The complete tool configurations, prompt templates, and skill-update settings are provided in the supplementary material.

\begin{figure*}[t]
\centering
\includegraphics[width=\linewidth]{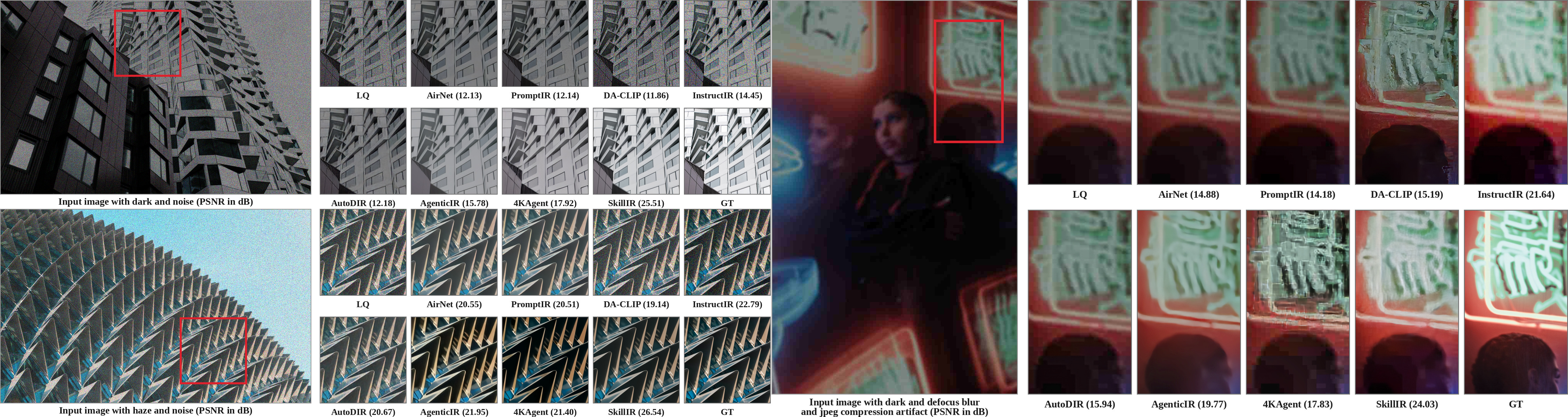}
\vspace{-1.0em}
\caption{Qualitative comparison on representative samples from MiO100 Groups A--C. Each row shows the degraded input and an enlarged view of the marked region. PSNR values are reported in dB.}
\label{fig:qualitative_results_hy}
\vspace{-1.0em}
\end{figure*}

\noindent\textbf{Datasets.}
For synthetic evaluation, we follow AgenticIR~\cite{zhu2025agenticir} and use 1,440 images from MiO100~\cite{kong2024mioir}, covering 16 mixed-degradation combinations across Groups A, B, and C. For real-world evaluation, we adopt the same 100-pair benchmark as SEAR~\cite{cui2026sear}, drawn from I-Haze~\cite{ancuti2018ihaze}, NH-Haze~\cite{ancuti2020nhhaze}, DRealSR~\cite{wei2020component}, RealSR~\cite{cai2019toward}, T-OLED~\cite{zhou2021image}, SIDD~\cite{abdelhamed2018high}, and LHP-Rain~\cite{guo2023sky}.

\subsection{Comparison with State-of-the-Art Methods}

\noindent\textbf{Results on MiO100.} Table~\ref{tab:mio100_all_merged_new} and Fig.~\ref{fig:qualitative_results_hy} present quantitative and qualitative comparisons across the three MiO100 groups. SkillIR consistently ranks first or second on the full-reference metrics. Compared with the strongest all-in-one baseline in PSNR, SkillIR improves by 1.55, 1.17, and 1.98\,dB on Groups A, B, and C, respectively, suggesting that adaptive tool orchestration is beneficial for handling mixed degradations beyond fixed all-in-one restoration models. Among agentic methods, SkillIR achieves the best LPIPS on all three groups and the best SSIM on Groups A and C. On Group B, although SEAR obtains 0.40\,dB higher PSNR and 0.0023 higher SSIM, SkillIR reduces LPIPS by 0.0235 and achieves higher MANIQA, CLIP-IQA, and MUSIQ scores, showing stronger perceptual quality despite slightly lower reconstruction-oriented metrics. On Group C, SkillIR nearly matches the best PSNR while achieving the best SSIM and LPIPS. 
Qualitatively, competing methods often retain residual degradations or oversmooth fine structures, whereas SkillIR better preserves building edges, repeated facade patterns, and neon-sign strokes.

\begin{figure*}[t]
\centering
\includegraphics[width=\linewidth]{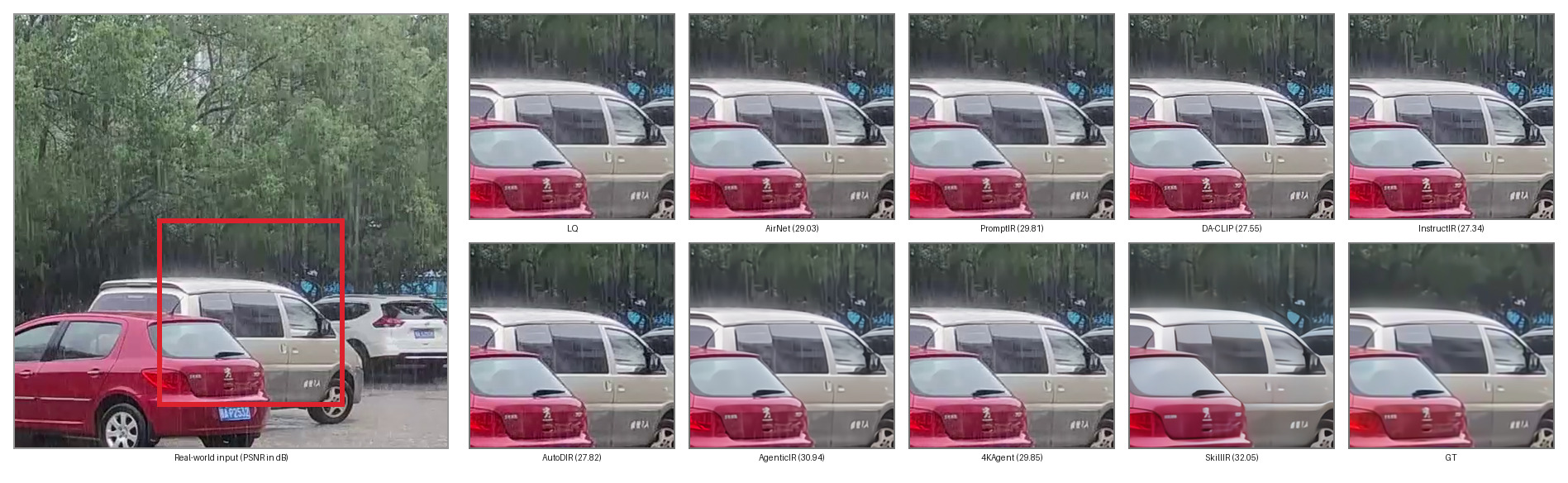}
\vspace{-2.0em}
\caption{Qualitative comparison on the paired real-world benchmark. 
}
\label{fig:skillir_showcase_real_world}
\vspace{-1.0em}
\end{figure*}

\noindent\textbf{Real-World Generalization.}
Table~\ref{tab:real_world_results} and Fig.~\ref{fig:skillir_showcase_real_world} report quantitative and qualitative results on the paired real-world benchmark. SkillIR achieves the best MANIQA, CLIP-IQA, and MUSIQ scores, together with the second-best PSNR and LPIPS. Compared with the strongest competing result for each no-reference metric, SkillIR improves MANIQA by 0.0441, CLIP-IQA by 0.0292, and MUSIQ by 1.12. Although InstructIR obtains higher PSNR and SSIM and SEAR achieves slightly better LPIPS, SkillIR delivers substantially stronger no-reference perceptual quality while maintaining competitive reconstruction fidelity. However, its lower SSIM shows that these perceptual gains do not translate uniformly to all fidelity-oriented metrics. Qualitatively, SkillIR removes complex real-world degradations while preserving clearer structures and introducing fewer visible artifacts than competing methods. These results suggest that restoration skills constructed from verified synthetic exploration trajectories remain effective when applied to unseen real-world images with different degradation characteristics.

\begin{table}[t]
\centering
\small
\setlength{\tabcolsep}{5.0pt}
\renewcommand{\arraystretch}{1.08}
\resizebox{\linewidth}{!}{
\begin{tabular}{@{}lcccccc@{}}
\toprule
Method
& PSNR$\uparrow$
& SSIM$\uparrow$
& LPIPS$\downarrow$
& MANIQA$\uparrow$
& CLIP-IQA$\uparrow$
& MUSIQ$\uparrow$ \\
\midrule
AirNet
& 23.3067 & 0.7471 & 0.4484 & 0.2356 & 0.3272 & 35.2690 \\
PromptIR
& 23.8647 & \underline{0.7542} & 0.4341 & 0.2405 & 0.3270 & 35.4811 \\
MiOIR
& 24.0043 & 0.7472 & 0.4239 & 0.2394 & 0.3507 & 36.8458 \\
DA-CLIP
& 23.7313 & 0.7437 & 0.4273 & 0.2514 & 0.3330 & 36.4113 \\
InstructIR
& \textbf{26.1477} & \textbf{0.7639}
& 0.4237 & 0.2556 & 0.3412 & 37.8236 \\
AutoDIR
& 20.7470 & 0.6640 & 0.4792 & 0.2466 & 0.3148 & 42.6789 \\
\midrule
AgenticIR
& 23.9280 & 0.7214 & 0.3730 & 0.3151 & 0.4512 & 52.9103 \\
4KAgent
& 23.6295 & 0.7242 & 0.3569
& \underline{0.3200} & 0.4513 & 54.4647 \\
SEAR
& 24.4078 & 0.7425 & \textbf{0.3371}
& 0.3174 & \underline{0.4519} & \underline{54.6686} \\
\textbf{SkillIR}
& \underline{24.5031} & 0.7203 & \underline{0.3459}
& \textbf{0.3641} & \textbf{0.4811} & \textbf{55.7843} \\
\bottomrule
\end{tabular}
}
\vspace{-0.5em}
\caption{Quantitative comparison on the paired real-world dataset.}
\label{tab:real_world_results}
\vspace{-1.0em}
\end{table}

\noindent\textbf{Efficiency Analysis.} Table~\ref{tab:efficiency} compares restoration quality and execution efficiency on Group B. SkillIR achieves the best LPIPS and the second-best PSNR while requiring the fewest restoration-tool calls. Compared with AgenticIR, the additional skill retrieval, transition verification, and residual re-perception increase end-to-end latency, but reduce the number of restoration-tool invocations. Compared with 4KAgent and SEAR, SkillIR requires both less inference time and fewer tool calls, demonstrating a favorable balance between restoration quality and execution efficiency.

\begin{table}[t]
\centering
\small
\setlength{\tabcolsep}{4.5pt}
\renewcommand{\arraystretch}{1.05}
\resizebox{0.75\linewidth}{!}{%
\begin{tabular}{@{}lcccc@{}}
\toprule
Method & PSNR$\uparrow$ & LPIPS$\downarrow$ & Time (min)$\downarrow$ & Tool Calls$\downarrow$ \\
\midrule
AgenticIR & 20.55 & 0.3072 & \textbf{1.09} & \underline{6.11} \\
4KAgent & 20.95 & 0.3017 & 2.55 & 8.26 \\
SEAR & \textbf{22.13} & \underline{0.2890} & 1.98 & 8.15 \\
SkillIR & \underline{21.73} & \textbf{0.2655} & \underline{1.63} & \textbf{5.84} \\
\bottomrule
\end{tabular}%
}
\vspace{-0.5em}
\caption{Efficiency comparison on Group B. Time denotes the average end-to-end inference time per image. Tool Calls counts all restoration-tool invocations, including candidates subsequently rejected by transition verification, while excluding MLLM and perception-model calls.}
\label{tab:efficiency}
\vspace{-1.5em}
\end{table}

\subsection{Ablation Study and Analysis}

We conduct ablation studies on Group B to evaluate skill initialization and evolution, episode-based construction, scene-aware retrieval, failure lessons, and transition verification. All variants use the same restoration toolbox, high-level controller, execution budget, and evaluation protocol.

\noindent\textbf{Component Ablation.}
As shown in Table~\ref{tab:module_ablation}, removing either dynamic or static skills degrades restoration performance, indicating that initialized skills and experience accumulated from completed rollouts provide complementary benefits. Without dynamic skills, PSNR decreases from 21.73 to 21.08\,dB and tool calls increase from 5.84 to 6.34, while removing static skills further reduces PSNR to 20.98\,dB. Replacing degradation-centered episode representation with full-trajectory retrieval also lowers restoration quality and increases tool calls from 5.84 to 7.88, showing that trajectory-level experience provides less localized guidance for evolving intermediate states. Removing scene context, failure lessons, or skill retrieval consistently degrades restoration quality, confirming that effective skill guidance relies on context-aware matching and the reuse of verified positive and negative experience. Among the verified-loop variants, removing transition verification causes the largest deterioration in PSNR, SSIM, LPIPS, and MUSIQ, demonstrating the importance of filtering harmful tool executions before commitment. In contrast, removing residual re-perception mainly affects perceptual quality, leading to notable drops in MANIQA and CLIP-IQA. Although this variant requires fewer tool calls, the reduction results from insufficient residual-state updates and premature termination of degradation processing rather than improved execution efficiency.
\begin{table}[t]
\centering
\small
\setlength{\tabcolsep}{3.5pt}
\renewcommand{\arraystretch}{0.95}
\resizebox{\linewidth}{!}{
\begin{tabular}{@{}lccccccc@{}}
\toprule
Setting & PSNR$\uparrow$ & SSIM$\uparrow$ & LPIPS$\downarrow$
& MANIQA$\uparrow$ & CLIP-IQA$\uparrow$ & MUSIQ$\uparrow$
& Calls$\downarrow$ \\
\midrule
Full SkillIR
& \textbf{21.73} & \textbf{0.7228} & \textbf{0.2655}
& \textbf{0.3675} & \textbf{0.5505} & \textbf{62.60}
& 5.84 \\

w/o Dynamic Skills
& 21.08 & 0.7038 & 0.2897 & 0.3526 & 0.5267 & 60.02 & 6.34 \\

w/o Static Skills
& 20.98 & 0.7008 & 0.2948 & 0.3503 & 0.5205 & 59.53 & 6.15 \\

\midrule
Full-Trajectory Retrieval
& 21.34 & 0.7118 & 0.2798 & 0.3594 & 0.5353 & 61.12 & 7.88 \\

w/o Scene Context
& 21.46 & 0.7147 & 0.2765 & 0.3609 & 0.5395 & 61.44 & 7.61 \\

w/o Failure Lessons
& \underline{21.50} & \underline{0.7163} & \underline{0.2739}
& \underline{0.3628} & \underline{0.5415} & \underline{61.73}
& 7.84 \\

w/o Skill Retrieval
& 21.18 & 0.7067 & 0.2869 & 0.3550 & 0.5285 & 60.36 & 7.06 \\

\midrule
w/o Transition Verification
& 20.87 & 0.6954 & 0.3036 & 0.3437 & 0.5141 & 58.51 & 6.72 \\

w/o Residual Re-perception
& 21.35 & 0.7086 & 0.2923 & 0.3298 & 0.4986 & 59.23 & 4.47 \\

\bottomrule
\end{tabular}
}
\vspace{-0.5em}
\caption{Component ablation on Group B. Full-Trajectory Retrieval
replaces degradation-centered episode representation with
trajectory-level experience retrieval while keeping the remaining
components unchanged.}
\label{tab:module_ablation}
\vspace{-2.0em}
\end{table}

\begin{figure}[t]
\centering
\includegraphics[width=\linewidth]{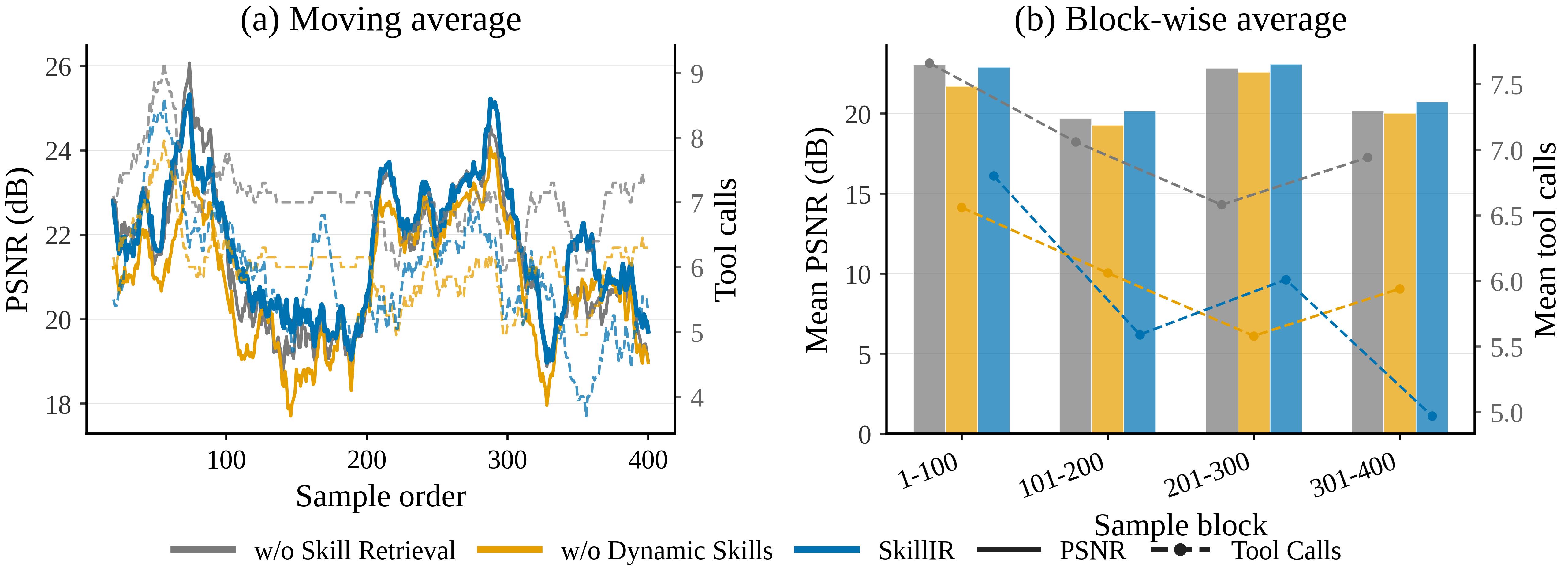}
\vspace{-1.2em}
\caption{Cross-sample skill evolution during deployment. 
Moving-average and block-wise trends of restoration quality and tool usage under different skill settings.}
\label{fig:skill_evolution_curve}
\vspace{-1.0em}
\end{figure}

\noindent\textbf{Cross-Sample Skill Evolution.}
Figure~\ref{fig:skill_evolution_curve} compares Full SkillIR with the \emph{w/o Dynamic Skills} and \emph{w/o Skill Retrieval} variants. Compared with these variants, SkillIR maintains competitive restoration quality while showing a decreasing trend in tool calls across successive samples. This trend indicates that verified experience from completed rollouts improves action selection for subsequent samples.

\begin{figure}[t]
\centering
\includegraphics[width=\linewidth]{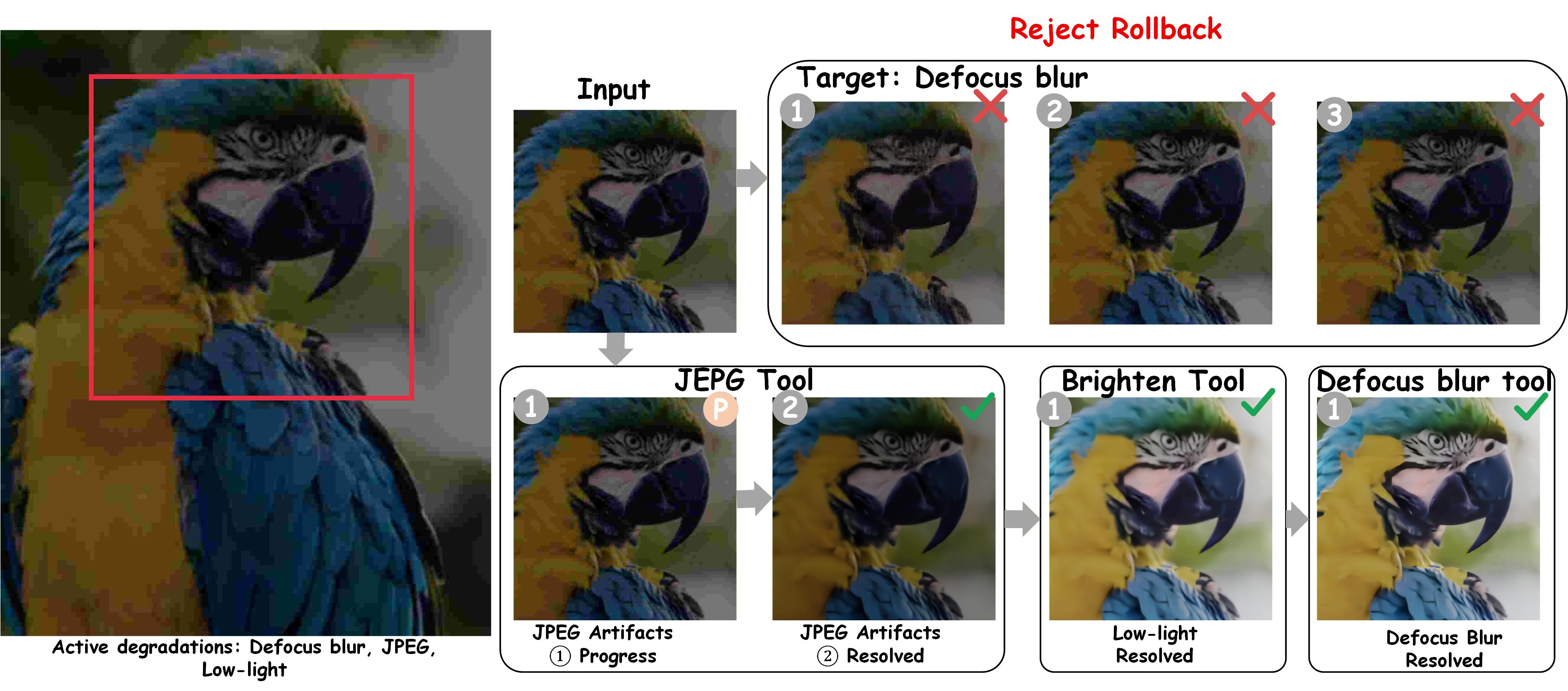}
\vspace{-1.5em}
\caption{A representative trajectory with candidate rejection, degradation switching, and residual re-perception.}
\label{fig:restoration_process}
\vspace{-1.0em}
\end{figure}

\noindent\textbf{Restoration Process.}
Figure~\ref{fig:restoration_process} shows a representative trajectory in which failed defocus-deblurring candidates are rejected, other active degradations are processed, and defocus blur is revisited after residual re-perception. The trajectory demonstrates how verification and re-perception enable rejection, degradation switching, and iterative restoration.

\section{Conclusion}

In this paper, we propose SkillIR, a skill-guided framework for agentic image restoration that learns reusable restoration knowledge from verified tool-use experience. By decomposing trajectories into degradation-centered episodes and organizing verified transitions into scene-aware skills with failure lessons, SkillIR enables state-adaptive action selection through a verified residual-state loop. Extensive experiments on synthetic and real-world multi-degradation benchmarks demonstrate that SkillIR improves restoration quality while reducing unnecessary tool executions. These results highlight the importance of representing restoration experience as verified local transitions for handling complex degradations with evolving intermediate states.
\bibliography{aaai2027}


\end{document}